%% file: paper.tex
\pdfoutput=1

\documentclass[11pt]{article}

\usepackage{emnlp2021}

\usepackage{times}
\usepackage{latexsym}
\usepackage[T1]{fontenc}
\usepackage[utf8]{inputenc}
\usepackage{microtype}

\usepackage{amsmath}
\usepackage{amssymb}

\title{Word-Level Coreference Resolution}

\author{Vladimir Dobrovolskii \\
  ABBYY / Moscow, Russia \\
  \texttt{v.dobrovolskii@abbyy.com} \\ }

\begin{document}
\maketitle
\begin{abstract}
Recent coreference resolution models rely heavily on span representations to find coreference links between word spans. As the number of spans is $O(n^2)$ in the length of text and the number of potential links is $O(n^4)$, various pruning techniques are necessary to make this approach computationally feasible. We propose instead to consider coreference links between individual words rather than word spans and then reconstruct the word spans. This reduces the complexity of the coreference model to $O(n^2)$ and allows it to consider all potential mentions without pruning any of them out. We also demonstrate that, with these changes, SpanBERT for coreference resolution will be significantly outperformed by RoBERTa. While being highly efficient, our model performs competitively with recent coreference resolution systems on the OntoNotes benchmark.
\end{abstract}

\section{Introduction}
Coreference resolution is used in various natural language processing pipelines, such as machine translation \citep{ohtani-etal-2019-context, miculicich-werlen-popescu-belis-2017-using}, question answering \citep{dhingra-etal-2018-neural} and text simplification \citep{wilkens-etal-2020-coreference}. As a building block of larger systems itself, it is crucial that coreference resolution models aim to not only achieve higher benchmark scores but be time and memory efficient as well.

Most recent coreference resolution models for English use the mention ranking approach: the highest ranking antecedent of each span is predicted to be coreferent to that span.
As this approach presents a serious computational challenge of $O(n^4)$ complexity in the document length, \citet{lee-etal-2017-end} keep only $M$ top-scoring spans, while \citet{lee-etal-2018-higher} extend this idea by keeping only top $K$ antecedents for each span based on an easily computable score. Their \textsc{c2f-coref} model, which showed 72.6 F1 on the OntoNotes 5.0 shared task \citep{pradhan-etal-2012-conll}, was later improved by \citet{joshi-etal-2020-spanbert}, who introduced SpanBERT to obtain better span representations (79.6 F1), and by \citet{xu-choi-2020-revealing}, who proposed a novel way of higher-order coreference resolution (80.2 F1).

The current state-of-the-art system built by \citet{wu-etal-2020-corefqa} uses a different approach, formulating the task as a machine reading comprehension problem. While they were able to achieve an impressive 83.1 F1 on the OntoNotes 5.0 shared task, their model is particularly computationally expensive as it requires a full transformer pass to score each span's antecedents.

To make coreference resolution models more compact and allow for their easier incorporation into larger pipelines, we propose to separate the task of coreference resolution from span extraction and to solve it on the word-level, lowering the complexity of the model to $O(n^2)$. The span extraction is to be performed separately only for those words that are found to be coreferent to some other words. An additional benefit of this approach is that there is no need for a complex span representation, which is usually concatenation-based. Overall this allows us to build a lightweight model that performs competitively with other mention ranking approaches and achieves 81.0 F1 on the OntoNotes benchmark.\footnote{Source code and models are available at \url{https://github.com/vdobrovolskii/wl-coref}}

\section{Related work}
\subsection{End-to-end coreference resolution}
The model proposed by \citet{lee-etal-2018-higher} aims to learn a probability distribution $P$ over all possible antecedent spans $Y$ for each span $i$:
\begin{equation}
    P(y_j) = \frac{e^{s(i, y_j)}}{\sum_{y' \in Y(i)}e^{s(i, y')}}
\end{equation}
Here $s(i, j)$ is the pairwise coreference score of spans $i$ and $j$, while $Y(i)$ contains spans to the left of $i$ and a special dummy antecedent $\epsilon$ with a fixed score $s(i, \epsilon) = 0$ for all $i$.

The pairwise coreference score $s(i, j)$ of spans $i$ and $j$ is a sum of the following scores:
\begin{enumerate}
    \item $s_m(i)$, whether $i$ is a mention;
    \item $s_m(j)$, whether $j$ is a mention;
    \item $s_c(i, j)$, coarse coreference score of $i$ and $j$;
    \item $s_a(i, j)$, fine coreference score of $i$ and $j$.
\end{enumerate}
They are calculated as follows:
\begin{align}
    s_m(i) &= \text{FFNN}_m(\mathbf{g}_i)    \\
    s_c(i, j) &= \mathbf{g}_i^\intercal \ \mathbf{W}_c \ \mathbf{g}_j  \\
    s_a(i, j) &= \text{FFNN}_a([\mathbf{g}_i, \mathbf{g}_j, \mathbf{g}_i \odot \mathbf{g}_j, \phi])
\end{align}
where $\mathbf{g}_i$ is the vector representation of span $i$ and $\phi$ is a vector of pairwise features, such as the distance between spans, whether they are from the same speaker, etc. The span representation $\mathbf{g}_i$ is initialized as a concatenation of contextual embeddings of the start and end tokens, the weighted sum of all the tokens in the span and a feature vector (learnable width and genre embeddings):
\begin{equation}
    \mathbf{g}_i = [\mathbf{x}_{\textsc{start}(i)}, \mathbf{x}_{\textsc{end}(i)}, \hat{\mathbf{x}}_i, \phi(i)]
\end{equation}
The weights to calculate $\hat{\mathbf{x}}_i$ are obtained using an attention mechanism \citep{bahdanau2014neural}.

The model also updated the span representations with the weighted sums of their antecedent representations to allow for cluster information to be available for a second iteration of calculating $\mathbf{S}_a$. However, \citet{xu-choi-2020-revealing} have demonstrated that its influence on the performance is negative to marginal.

\subsection{Coreference resolution without span representations}
Recently, \citet{kirstain-etal-2021-no-span} have proposed a modification of the mention ranking model which does not use span representations. Instead, they compute the start and end representations of each subtoken in the sequence:
\begin{align}
    \mathbf{X}^s &= \textsc{ReLU}(\mathbf{W}^s \cdot \mathbf{X}) \\
    \mathbf{X}^e &= \textsc{ReLU}(\mathbf{W}^e \cdot \mathbf{X})
\end{align}
The resulting vectors are used to compute mention and antecedent scores for each span without the need to construct explicit span representations. This approach performs competitively with other mention ranking approaches, while being more memory efficient. However, the theoretical size of the antecedent score matrix is still $O(n^4)$, so the authors need to prune the resulting mentions. In our paper, we present an approach that does not exceed the quadratic complexity without the need for mention pruning, while retaining the comparable performance.

\section{Model}
\subsection{Token representation}
After obtaining contextual embeddings of all the subtokens of a text, we compute token representations $\mathbf{T}$ as weighted sums of their respective subtokens. The weights are obtained by applying the softmax function to the raw scores of subtokens of each token. The scores are calculated by multiplying the matrix of subtoken embeddings $\mathbf{X}$ by a matrix of learnable weights $\mathbf{W}_a$:
\begin{equation}
    \mathbf{A} = \mathbf{W}_a \cdot \mathbf{X}
\end{equation}
No mention score is computed for the resulting token representations and none of them is subsequently pruned out.

\subsection{Coarse-to-fine antecedent pruning}
Following \citet{lee-etal-2018-higher} we first use a bilinear scoring function to compute $k$ most likely antecedents for each token. This is useful to further reduce the computational complexity of the model.
\begin{equation}
    \textbf{S}_c = \mathbf{T} \cdot \mathbf{W}_c \cdot \mathbf{T}^\intercal
\end{equation}
Then we build a pair matrix of $n \times k$ pairs, where each pair is represented as a concatenation of the two token embeddings, their element-wise product and feature embeddings (distance, genre and same/different speaker embeddings). The fine antecedent score is obtained with a feed-forward neural network:
\begin{equation}
    s_a(i, j) = \textsc{FFNN}_a([\mathbf{T}_i, \mathbf{T}_j, \mathbf{T}_i \odot \mathbf{T}_j, \phi])
\end{equation}
The resulting coreference score is defined as the sum of the two scores:
\begin{equation}
    s(i, j) = s_c(i, j) + s_a(i, j)
\end{equation}
The candidate antecedent with the highest positive score is assumed to be the predicted antecedent of each token. If there are no candidates with positive coreference scores, the token is concluded to have no antecedents.

Our model does not use higher-order inference, as \citet{xu-choi-2020-revealing} have shown that it has a marginal impact on the performance. It is also computationally expensive since it repeats the most memory intensive part of the computation during each iteration.

\subsection{Span extraction}
The tokens that are found to be coreferent to some other tokens are further passed to the span extraction module. For each token, the module reconstructs the span by predicting the most probable start and end tokens in the same sentence. To reconstruct a span headed by a token, all tokens in the same sentence are concatenated with this head token and then passed through a feed-forward neural network followed by a convolution block with two output channels (for start and end scores) and a kernel size of three. The intuition captured by this approach is that the best span boundary is located between a token that is likely to be in the span and a token that is unlikely to be in the span. During inference, tokens to the right of the head token are not considered as potential start tokens and tokens to the left are not considered as potential end tokens.

\subsection{Training}
To train our model, we first need to transform the OntoNotes 5.0 training dataset to link individual words rather than word spans. To do that, we use the syntax information available in the dataset to reduce each span to its syntactic head. We define a span's head as the only word in the span that depends on a word outside the span or is the head of the sentence. If the number of such words is not one, we choose the rightmost word as the span's head. This allows us to build two training datasets: 1) word coreference dataset to train the coreference module and 2) word-to-span dataset to train the span extraction module.

Following \citet{lee-etal-2017-end}, the coreference module uses negative log marginal likelihood as its base loss function, since the exact antecedents of gold spans are unknown and only the final gold clustering is available:
\begin{equation}
    L_{\textsc{nlml}} = -\log \prod_{i = 0}^{N} \ \sum_{y' \in Y(i) \cap \textsc{gold}(i)} P(y')
\end{equation}
We propose to use binary cross-entropy as an additional regularization factor:
\begin{equation}
    L_\textsc{coref} = L_\textsc{nlml} + \alpha L_\textsc{bce}
\end{equation}
This encourages the model to output higher coreference scores for all coreferent mentions, as the pairs are classified independently from each other. We use the value of $\alpha = 0.5$ to prioritize NLML loss over BCE loss.

The span extraction module is trained using cross-entropy loss over the start and end scores of all words in the same sentence.

We jointly train the coreference and span extraction modules by summing their losses.

\section{Experiments}
\input{table_test}

We use the OntoNotes 5.0 test dataset to evaluate our model and compare its performance with the results reported by authors of other mention ranking models. The development portion is used to reason about the importance of the decisions in our model's architecture.

\subsection{Experimental setup} We implement our model with PyTorch framework \citep{pytorch}. We use the Hugging Face Transformers \citep{wolf-etal-2020-transformers} implementations of BERT \citep{devlin-etal-2019-bert}, RoBERTa \citep{liu2019roberta}, SpanBERT \citep{joshi-etal-2020-spanbert} and Longformer \citep{longformer}. All the models were used in their \emph{large} variants.
We also replicate the span-level model by \citet{joshi-etal-2020-spanbert} to use it as a baseline for comparison.

We do not perform hyperparameter tuning and mostly use the same hyperparameters as the ones used in the \emph{independent} model by \citet{joshi-etal-2020-spanbert}, except that we also linearly decay the learning rates.

Our models were trained for 20 epochs on a 48GB nVidia Quadro RTX 8000. The training took 5 hours (except for the Longformer-based model, which needed 17 hours). To evaluate the final model one will need 4 GB of GPU memory.

\subsection{Selecting the best model} The results in Table \ref{table_dev} demonstrate that the word-level SpanBERT-based model performs better than its span-level counterpart (\textsc{joshi-replica}), despite being much more computationally efficient.
The replacement of SpanBERT by RoBERTa or Longformer yields even higher scores for word-level models, though RoBERTa does not improve the performance of the span-level model.

The addition of binary cross-entropy as an extra regularization factor has a slight positive effect on the performance. As it introduces zero overhead on inference time, we keep it in our final model.

All the underlying transformer models (with BERT being the exception) have reached approximately the same quality of span extraction. While it allows the approach to compete with span-level coreference models, there is still room for improvement, which can be seen from the gap between "pure" word-level coreference scores and span-level scores.
\input{table_dev}

\subsection{Efficiency} As one can see from the Table \ref{table_speed_theoretical}, the word-level approach needs to consider 14x fewer mentions and 222x fewer mention pairs than the span-level approach. This allows us to keep all the potential mentions while span-level approaches have to rely on pruning techniques, like scoring the mentions and selecting the top-scoring portion of them for further consideration. Such pruning is a compromise between efficiency and accuracy, so removing the necessity for it positively influences both. Moreover, our representation of a mention does not require concatenating start, end and content vectors, which additionally reduces the memory requirements of the model.

\input{table_speed_theoretical}
\input{table_speed_measurements}

Table \ref{table_speed_measurements} contains actual efficiency measurements of our baseline (\textsc{joshi-replica}) before and after disabling higher-order inference and switching to word-level coreference resolution. While the baseline relies on aggressive mention pruning and keeps only top $\lambda n$ mentions, where $\lambda = 0.4$ and $n$ is the number of words in the input text, it still requires more time and memory than its word-level counterpart, which considers all possible mentions.

As a reference, we also provide time and memory measurements for other mention ranking models. However, they should not be directly compared, as there are other factors influencing the running time, such as the choice of a framework, the degree of mention pruning, and code quality.

\subsection{Test results} Table \ref{table_test} demonstrates that our model performs competitively with other mention ranking approaches. It can be seen that it has a higher recall than span-level models, because it does not need any mention pruning techniques to be computationally feasible and considers all the potential antecedents for each token in the text.

\section{Conclusion}
We introduce a word-level coreference model that, while being more efficient, performs competitively with recent systems. This allows for easier incorporation of coreference models into larger natural language processing pipelines. In addition, the separation of coreference resolution and span extraction tasks makes the model's results more interpretable.
If necessary, the span prediction module can be replaced by a syntax-aware system to account for non-continuous spans.
Such task separation also makes the coreference module independent from the way coreferent spans are marked up in the training dataset. This will potentially simplify using various data sources to build robust coreference resolution systems.

\bibliography{anthology,custom}
\bibliographystyle{acl_natbib}




\end{document}

%% file: table_test.tex
\begin{table*}
\centering
\begin{tabular}{lcccccccccc}
\hline
& \multicolumn{3}{c}{\textbf{MUC}} & \multicolumn{3}{c}{\textbf{B$^3$}} & \multicolumn{3}{c}{\textbf{CEAF$_{\phi4}$}} & \\
& P & R & F1 & P & R & F1 & P & R & F1 & Mean F1 \\ \hline

\citet{joshi-etal-2020-spanbert} &
85.8 & 84.8 & 85.3 &
78.3 & 77.9 & 78.1 &
76.4 & 74.2 & 75.3 &
79.6 \\

\textsc{joshi-replica} &
85.2 & 85.5 & 85.4 &
78.2 & 78.7 & 78.4 &
75.4 & 75.2 & 75.3 &
79.7 \\

\citet{xu-choi-2020-revealing} &
85.9 & 85.5 & 85.7 &
79.0 & 78.9 & 79.0 &
76.7 & 75.2 & 75.9 &
80.2 \\

\citet{kirstain-etal-2021-no-span} &
\textbf{86.5} & 85.1 & 85.8 &
\textbf{80.3} & 77.9 & 79.1 &
\textbf{76.8} & 75.4 & 76.1 &
80.3 \\

\hline

wl-coref + RoBERTa &
84.9 & \textbf{87.9} & \textbf{86.3} &
77.4 & \textbf{82.6} & \textbf{79.9} &
76.1 & \textbf{77.1} & \textbf{76.6} &
\textbf{81.0} \\

\hline
\end{tabular}
\caption{\label{table_test} Best results on the OntoNotes 5.0 test dataset. All the scores have been obtained using the official CoNLL-2012 scorer \citep{pradhan-etal-2014-scoring} and rounded to 1 decimal place. MUC is the metric proposed by \citet{vilain-etal-1995-model}, B$^3$ was introduced by \citet{Bagga98algorithmsfor} and CEAF$_{\phi4}$ was designed by \citet{luo-2005-coreference}. The usage of mean F1 score as the aggregate metric was suggested by \citet{pradhan-etal-2012-conll}. }
\end{table*}

%% file: table_dev.tex
\begin{table} 
    \centering
    \begin{tabular}{lccc}
        \hline
        & \textbf{WL F1} & \textbf{SA} & \textbf{SL F1} \\ \hline

        wl + RoBERTa                  & 83.11 & 97.16 & 80.72 \\
        \hspace{5pt} -BCE           & 83.05 & 97.11 & 80.60 \\
        wl + SpanBERT                 & 82.52 & 97.13 & 80.14 \\
        \hspace{5pt} -BCE           & 82.32 & 97.10 & 79.99 \\
        wl + BERT                     & 77.55 & 96.20 & 74.80 \\
        wl + Longformer               & 82.98 & 97.14 & 80.56 \\

        \hline

        \textsc{joshi-replica}      & n/a & n/a & 79.74 \\
        \hspace{5pt} +RoBERTa       & n/a & n/a & 78.65 \\

        \hline
    \end{tabular}
    \caption{Model comparisons on the OntoNotes 5.0 development dataset (best out of 20 epochs). \textbf{WL F1} means word-level CoNLL-2012 F1 score, i.e.\ the coreference metric on the word-level dataset; \textbf{SA} is the span extraction accuracy or the percentage of correctly predicted spans; \textbf{SL F1} is the span-level CoNLL-2012 F1 score, the basic coreference metric.}
    \label{table_dev}
\end{table}

%% file: table_speed_theoretical.tex
\begin{table} 
    \centering
    \begin{tabular}{lrrr}
        \hline
        & \textbf{Mentions} & \textbf{Mention pairs} & \textbf{SBC} \\ \hline

        WL  &   163,104 &     62,803,841    & 475,208   \\
        SL  & 2,263,299 & 13,970,813,822    & n/a       \\
        
        \hline
    \end{tabular}
    \caption{The total number of mentions and mention pairs under the word-level and span-level approaches on the OntoNotes 5.0 development dataset. The spans do not cross sentence boundaries. For each mention, only mentions to its left are counted as mention pairs. The total number of span boundary candidates (\textbf{SBC}) that need to be considered to predict spans is negligible compared to the number of mention pairs. }
    \label{table_speed_theoretical}
\end{table}

%% file: table_speed_measurements.tex
\begin{table} 
    \centering
    \begin{tabular}{lrr}
        \hline
        & \textbf{Time} & \textbf{Memory} \\ \hline

        \textsc{joshi-replica}      & 95    & 7.4   \\
        \hspace{5pt} -HOI           & 88    & 7.4   \\
        \hspace{15pt} +wl-coref     & 32    & 2.9   \\
        
        \hline
        
        \citet{joshi-etal-2020-spanbert}    & 85    & 13.3  \\
        \citet{xu-choi-2020-revealing} CM   & 92    & 12.4  \\
        \hspace{5pt} -HOI                   & 52    & 12.4  \\
        \citet{kirstain-etal-2021-no-span}  & 44    & 4.0   \\
        
        \hline
    \end{tabular}
    \caption{Inference time (sec) and peak GPU memory usage (GiB) on the OntoNotes 5.0 development dataset. The measurements were made using nVidia Quadro RTX 8000. }
    \label{table_speed_measurements}
\end{table}